\documentclass{article}
\usepackage[table,xcdraw]{xcolor}
\usepackage[utf8]{inputenc}
\usepackage{graphicx}
\usepackage{mathtools}
\usepackage{float}
\usepackage{url}
\usepackage{algorithm}
\usepackage{algpseudocode}
\usepackage{tikz,forest}
\usepackage{listings}
\usetikzlibrary{arrows.meta}
\usepackage[skins, breakable]{tcolorbox}
\usepackage{amsthm}
\usepackage{amssymb}
\usepackage{todonotes}
\usepackage{authblk}
\begin{document}

\title{Comparing decision mining approaches with regard to the meaningfulness of their results}

\author[1]{Beate Scheibel} 
\author[2]{Stefanie Rinderle-Ma} 
\affil[1]{Research Group Workflow Systems and Technology,
Faculty of Computer Science,
University of Vienna,
beate.scheibel@univie.ac.at }
\affil[2]{
Chair of Information Systems and Business Process Management\linebreak
Department of Informatics\linebreak
Technical University of Munich, Germany\linebreak
stefanie.rinderle-ma@tum.de
}
\date{}
\maketitle
\begin{abstract}

Decisions and the underlying rules are indispensable for driving process execution during runtime, i.e., for routing process instances at alternative branches based on the values of process data. 
Decision rules can comprise unary data conditions, e.g., age $>$ 40, binary data conditions where the relation between two or more variables is relevant, e.g. temperature1 $<$ temperature2, and more complex conditions that refer to, for example, parts of a medical image. Decision discovery aims at automatically deriving decision rules from process event logs. Existing approaches focus on the discovery of unary, or in some instances binary data conditions. The discovered decision rules are usually evaluated using accuracy, but not with regards to their semantics and meaningfulness, although this is crucial for validation and the subsequent implementation/adaptation of the decision rules. Hence, this paper compares three decision mining approaches, i.e., two existing ones and one newly described approach, with respect to the meaningfulness of their results. For comparison, we use one synthetic data set for a realistic manufacturing case and the two real-world BPIC 2017/2020 logs. The discovered rules are discussed with regards to their semantics and meaningfulness.
\end{abstract}

\section{Introduction}
\label{sect:intro}

Process mining enables process discovery, conformance checking, and process enhancement \cite{aalst_process_2016}. One important aspect of process discovery is the derivation of decision points and the corresponding decision rules based on event logs \cite{leoni_decision_2018}. 
The discovered decision rules are usually of the form $v(ariable)\ op(erator)\ c(onstant)$, i.e., a variable compared to a constant value (referred to as \textsl{unary data condition}), e.g., \textsl{temperature below 50°}. However, real-world processes often include more complex decision rules \cite{dunkl_method_2015}. Then, restricting discovery to unary conditions can lead to semantically incorrect decision rules that perform well because of overfitting, but do not appropriately depict the underlying decision logic. 
The first step towards decision rules reflecting real-world needs is to consider \textsl{binary data conditions}.
These conditions involve two or more variables that are in a specific relation to each other,
e.g., $t_1$ $<$ $t_2$ where $t_1$ and $t_2$ are variables and their values are written during process runtime. 
Approaches such as \texttt{BranchMiner} 
\cite{de_leoni_discovering_2013} enable the discovery of binary data conditions. However, again the focus is on accuracy of the results; their semantics is not considered despite the increased complexity of the rules. 

Considering the semantics and meaningfulness of decision mining results is especially important with respect to \textsl{transparency} and \textsl{explainability} which are requirements for process mining \cite{leewis_future_2020}.
Another challenge is \textsl{validity}. A distinction between internal and external validity is made, where internal validity can be loosely translated to the accuracy, whereas external validity refers to whether the result can be generalized \cite{leewis_future_2020}. This translates to decision mining as well. The discovered decision rules should be accurate, generalizable as well as transparent in the sense that they depict the underlying logic.
Transparency and explainability are current challenges for machine learning in general \cite{guidotti_survey_2018}, and for decision mining in particular. Decision rules should be accurate as well as contain meaningful rules. A first step is the ability to discover more complex decision rules. However, this might also lead to an increased number of discovered rules, which in turn amplifies the need to check for meaningfulness in addition to accuracy. 

This paper compares three decision mining approaches that enable the discovery of unary and binary data conditions with respect to the semantics and meaningfulness of the results. The first approach features a basic decision tree implementation. The second approach is \texttt{BranchMiner} (BM) 
\cite{de_leoni_discovering_2013} which is available as a compiled tool without source code or detailed documentation. Therefore, a third approach -- the Extended Decision Tree (EDT) -- is provided. It builds upon the concepts of the basic tree implementation and integrates ideas from \cite{de_leoni_discovering_2013} in order to enable dynamic testing and ensure transparency and comprehensibility of the rules. This approach is currently restricted to binary data conditions with comparison operators, which is sufficient for the purposes of this paper. The three approaches are compared based on two real life datasets and one synthetic dataset. Based on the observations from the comparison we provide recommendations for the interpretation and validation of decision mining results. 

Section \ref{sect:fundamentals} presents a running example from the manufacturing domain. The decision mining approaches to be compared will be described in Sect. \ref{sect:approach}. The results are described in Sect. \ref{sect:eval} and analyzed in Sect. \ref{sect:discussion}. In Sect. \ref{sect:rw} the related work is discussed and Sect. \ref{sect:conclusion} provides a conclusion.

\section{Running Example: Measuring Process in Manufacturing}
\label{sect:fundamentals}

\begin{figure*}[h]
    \centering
    \includegraphics[scale=0.20]{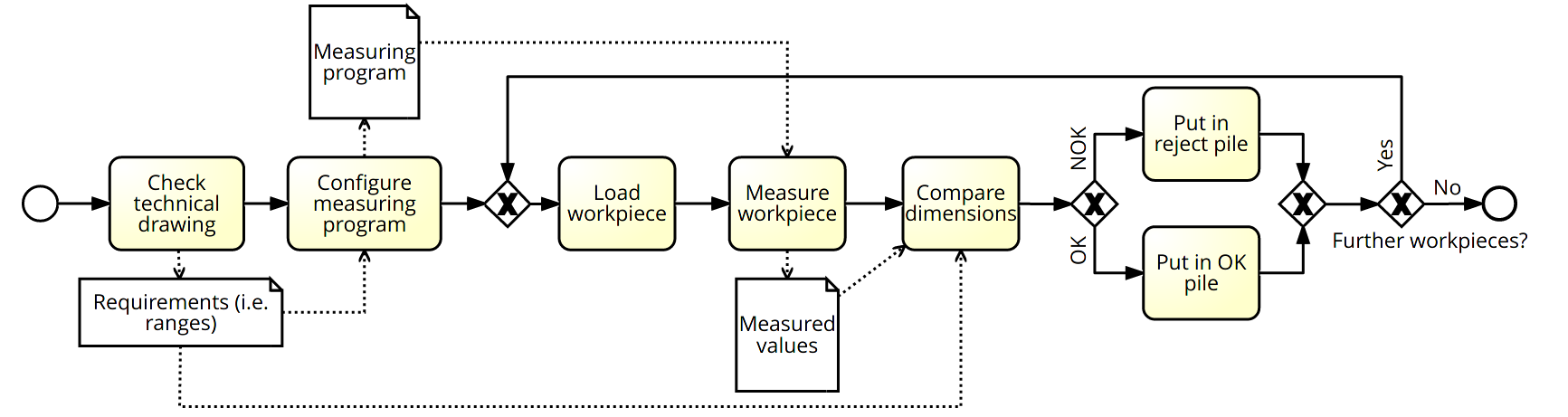}
    \caption{BPMN model of measuring process (modeled using Signavio\textsuperscript{\textcopyright}).}
    \label{fig:process_before}
\end{figure*}

This section describes a use case from the manufacturing domain to illustrate the challenges described in Sect. \ref{sect:intro}. In the manufacturing domain producing workpieces and checking their quality is a core task. Workpieces are manufactured manually or using different machines and subsequently checked if they adhere to the conditions (nominal value and tolerances) specified in the technical drawing or CAD model. Measuring can be done manually or using a coordinate measuring machine. If a measuring machine is used, the measuring program specifies which features have to be measured for a particular workpiece. This program has to be configured manually and is based on the technical drawing. After the workpiece is measured, the measured values have to be checked to make sure they adhere to the specified conditions. Then it is either placed on the scrap pile or on a tray for further use and the process can start again. If the measuring is done manually, some measuring tool, e.g., a calliper will be used. The rest of the process is performed the same way. The corresponding process model is depicted in Fig. \ref{fig:process_before}. 

During runtime the events reflecting the process execution are stored in a \textsl{process event log}. Listing \ref{logfile} shows an excerpt of a process event log for the process model in Fig. \ref{fig:process_before}. The uuid identifies the trace of a particular process instance. The logged events reflect the execution of process tasks \texttt{Prepare attribute list} and \texttt{Measure workpiece} with their timestamps and associated data elements.

{\footnotesize
\begin{lstlisting}[numbers=left, xleftmargin=7mm, caption= Example log in XES format, label=logfile]
<trace>
 <int key='uuid' value='6775'/>
 <event>
   <string key='concept:name' 
        value='Check technical drawing'/>
   <string key='ranges' 
        value='[[20,25],[10,15],[30,35]]'/>
   <date key='time:timestamp' 
        value='2021-03-30T09:21:30.423+02:00'/>
 </event>
 ...
 <event>
   <string key='concept:name' 
        value='Measure workpiece'/>
   <string key='measured_values' 
        value='[24,12,31]'/>
   <date key='time:timestamp' 
        value='2021-03-30T09:21:30.460+02:00'/>
  </event>
  ...
</trace>
\end{lstlisting}
}

This process includes one decision point i.e., \texttt{Dimensions in tolerance?}, which can be answered by 'Yes' or 'No' and the respective paths lead to either \texttt{Put in OK Pile} or \texttt{Put in NOK Pile}. However, the underlying decision rule is more complex (i.e., 'all measured features have to be in the specified tolerance zone') and can only be obtained by discussing the process with a domain expert. When performing this process manually, the decision is made by the employee, who documents the measured values either digitally or on a piece of paper. When automating the entire process, the \texttt{attribute list}, the list of all essential features with the respective tolerances is stored digitally. When the feature is measured, it can be automatically checked if it lies within the tolerance range. This can be hard-coded as part of the process. However, it is more challenging to automatically discover the underlying decision rule from the event log. 
Table \ref{table:og} can be obtained by merging the event data from one trace into one row. It shows the identifier (the uuid), the tolerance ranges, the measured values (abbreviated as 'meas' in the following), as well as the result, i.e., if the workpiece was specified as 'OK' or 'NOK'. All list values have to be split up into single values, in order to be used as input for the three decision mining approaches.

\begin{table}[H]
\centering
\caption{Data for running example 1.}
\label{table:og}
\begin{tabular}{llll}
uuid & ranges & meas & result  \\
0001 & [(20,25),(10,15),(30,35)] & [24,12,31] & OK   \\
0002 & [(20,25),(10,15),(30,35)] & [23,13,33] & OK   \\
0003 & [(20,25),(10,15),(30,35)] & [24,10,37] & NOK  
\end{tabular}
\end{table}

\section{Discovering Decision Rules with Extended Data Conditions}
\label{sect:approach}

In general, algorithms for decision mining have to fulfill a major requirement: the resulting decision rules have to be human readable. Therefore decision trees are typically used for decision mining instead of black-box approaches such as neural networks  \cite{leoni_decision_2018}.
In the following, we illustrate the challenges of the basic decision mining approach based on the running example (cf. Sect. \ref{sect:fundamentals}), before the BranchMiner \cite{de_leoni_discovering_2013} is described, and the Extended Decision Tree is introduced.

\subsection{Basic Decision Tree - BDT}
\begin{figure}[]
 \centering 
 { \footnotesize
\begin{forest} 
[$meas0 > 21$
    [NOK, edge label={node[midway,left]{false}}  ]   
        [$meas1 < 19$, edge label={node[midway,right]{true}}  
            [NOK]
            [$meas2 > 31$
                [NOK]
                [$meas2 < 69$
                    [NOK]
                    [$meas1 > 9$
                        [NOK]
                        [$meas0 < 79$
                        [NOK]
                        [OK]
                        ]
                ]
            ]]]   
] 
\end{forest}
}
\caption{Decision tree generated by basic approach.}
\label{tree:traditional}
\end{figure}
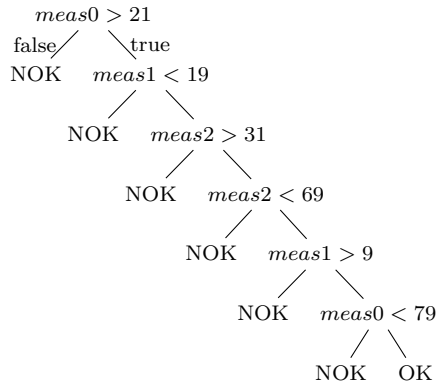

Most approaches for discovering decision rules use decision trees. Well-known algorithms include CART \cite{breiman_classification_1984} and C4.5 \cite{quinlan_j_r_c45_1993}. Decisions are treated as a classification problem. In the running example, the decision tree can be used to classify a certain workpiece in 'OK' or 'NOK'. The possible input formats include numerical data and, depending on the implementation, categorical data. 
For the running example, a CART implementation \cite{pedregosa_scikit-learn_2011} generates the decision tree in Fig. \ref{tree:traditional}, after the lists have been transformed into single valued variables. The classification uses unary data conditions, as current decision tree implementations are not able to discover binary data conditions. The basic approach works in many cases and can achieve high accuracy. However, valuable semantic information is lost in the process as we do not know the exact ranges and the underlying relations which can lead to misleading rules and lower accuracy if the input data changes. For example, the learning set could contain only a slight variation in values. Thus the learned conditions contain smaller ranges than stated in the drawing. This is the case in the running example, as well. If the conditions contained in Fig. \ref{tree:traditional} are aggregated, the tolerance ranges are (22,78),(10,18),(32,69) instead of the ranges specified in the drawing: (20,80),(10,20),(30,70). Here, the discovered rule contains no semantic information and is slightly inaccurate.

\subsection{BranchMiner - BM}
BranchMiner \cite{de_leoni_discovering_2013} uses Daikon \cite{ernst_daikon_2007}, a tool for discovering invariants in Software, to detect invariants in event logs. Invariants refer to data conditions that are true at one point in the process, e.g., before Event E takes place Variable v has to have a specific value. It is able to discover unary as well as binary data conditions and even linear relationships e.g. $v1*10<v2$. 
Binary conditions are discovered using \textsl{latent variables}. Latent variables are additional variables that are generated by combining all variables with each other using arithmetic operators ($+,-,*,/$). After these invariants are discovered, a decision tree, here the Weka \cite{eibe_weka_2016} implementation, is used to generate decision rules. The tool and some log files for testing are available online, but no detailed documentation or source code could be found. 

{\footnotesize
\begin{table*}[]
\caption{Running example with latent variables.}
\label{table:add_features}
\begin{tabular}{lllllll
>{\columncolor[HTML]{FFFC9E}}l 
>{\columncolor[HTML]{FFFC9E}}l ll}
uuid &
  range1 &
  range5 &
  meas0 &
  meas2 &
  ... &
  \begin{tabular}[c]{@{}l@{}}range1\\ $>=$\\ range5\end{tabular} &
  \begin{tabular}[c]{@{}l@{}}range1\\ $>=$\\ meas0\end{tabular} &
  \begin{tabular}[c]{@{}l@{}}meas2\\ $<=$\\ range5\end{tabular} &
  .... &
  res \\
0001 & 25 & 35 & 24 & 31 & ... & false & true & true  & ... & OK  \\
0002 & 25 & 35 & 23 & 33 & ... & false & true & true  & ... & OK  \\
0003 & 25 & 35 & 24 & 37 & ... & false & true & false & ... & NOK
\end{tabular}
\end{table*} 
}

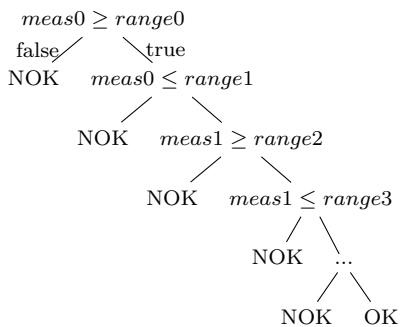
\begin{figure}[]
\centering 
{ \footnotesize
\begin{forest} 
[$meas0 \geq range0$
    [NOK, edge label={node[midway,left]{false}}]   
        [$meas0 \leq range1$, edge label={node[midway,right]{true}}
            [NOK]
            [$meas1 \geq range2$
            [NOK]
            [$meas1 \leq range3$
            [NOK]
            [...
            [NOK]
            [OK]
            ]]]]]     
] 
\end{forest}
}
\caption{Decision tree including binary data conditions.}
\label{tree:new}
\vspace{-5mm}
\end{figure}

\subsection{Extended Decision Tree - EDT}
BDT is not able to discover binary decision rules, whereas BM detects complex decision rules, but cannot be modified or adapted. Therefore we propose EDT, that integrates the concept of latent variables into BDT in order to ensure transparency, comprehensibility as well as easy modification of the source code.

\cite{blanco-justicia_machine_2020} emphasize the importance of comprehensibility, i.e., explanations that are understandable to humans and the role decision trees can play to provide more comprehensive results. EDT allows to generate a decision tree alongside with the discovered decision rules, similar to Fig. \ref{tree:new}. The decision tree can be useful to understand where a specific condition is used, hence contributing to comprehensibility of the results. In addition, the available source code ensures easier testing and adaptation. However, arithmetic operators or linear relationships are not included as part of this study, i.e., EDT is only able to discover binary data conditions with comparison operators. EDT will be adapted to include more complex relationships in future work. 

The first step of EDT is to convert the log files into table format. Additionally, as for the running example, the lists of values have to be split into single values, which was also the case for BDT and BM. Furthermore the result column has to be defined. Numeric variables are used as input variables. As for BM, latent variables enable the discovery of binary conditions. Latent variables are generated by taking two variables and comparing them using an operator $\otimes$, with $\otimes \in$ \{$<, >, \leq, \geq, \neq, =$\}. The resulting logical value is stored in a new variable.

This is done for all variables and all comparison operators. The result for the running example can be partially seen in Tab. \ref{table:add_features}. The data element \textsl{ranges} from Tab. \ref{table:og} was split into singular values (range0-range5), similar \textsl{measured\_values} was split into meas0-meas2, this was also done for BDT. Three latent variables are shown, i.e., range1$>=$range5, range$>=$meas0 and meas2$<=$range5. The last column shows the result i.e., if the workpiece was classified as OK or NOK. The yellow columns highlight potentially relevant variables.

Data conditions are learned using an existing decision tree implementation, i.e., sklearn  \cite{pedregosa_scikit-learn_2011}. Using an already tested and optimized implementation allows acceptable runtimes even for large datasets. This is important as the number of elements increases with the generation of new elements. After the decision tree is generated, we can access the different paths for the different result classes e.g. the conditions that have to be met by the workpiece to be qualified as 'OK'. The last step is to create a conjunctive decision rule by joining the conditions. This can be done for each outcome class. 

\section{Evaluation}
\label{sect:eval}

The prototypical implementations of BDT and EDT can be found here\footnote{\url{https://github.com/bscheibel/edt}}. 

We aim at comparing the decision rules discovered by BDT, BM, and EDT (cf. Sect. \ref{sect:approach}). As the decision rules might contain binary data conditions, the data sets used in the evaluation must involve at least two numerical variables and a result variable. 
The first data set contains synthetic data for the running example (cf. Sect. \ref{sect:fundamentals}) and is available online$^1$.
In addition, we use two data sets from the Business Process Mining Challenge (BPIC), i.e., from 2017 and 2020. BPIC2017 and BPIC2020 include XES log files, which have to be transformed into 'dataframes' for BDT and EDT and to 'cpn' files for the BM approach. BM was executed with the option 'latent variables' enabled. 
The data sets are split into training and test sets (90\% training, 10\% test), which can be used to determine the accuracy of the results. To calculate the accuracy, the following formula is used: 
\[Accuracy := \frac{Number\ of\ correctly\ classified\ instances}{Total\ number\ of\ instances}\]

Here, instances refers to actual process instances, i.e., the different cases, identified by some identifier, for example the 'uuid' in the running example. The term 'classified' refers to the result, i.e., if the result can either be 'OK' or 'NOK', the test set instances are classified as one of these according to the discovered decision rule, and then compared to the actual result to check if the instance was correctly classified.

In addition to accuracy, the recall and precision value for the discovered rules are calculated, based on the definition by \cite{ting_precision_2010}.
To calculate these values, a ground truth has to be known, which refers to the conditions that are part of the actual underlying decision rule. As the ground truth is only defined for the running example, the other rules are estimated, based on the information available.

Recall refers to the ratio of discovered relevant conditions to the number of conditions in the ground truth.  'Relevant' in this context means conditions that are also part of the ground truth decision rule. 

 \[Recall := \frac{Total\ number\ of\ discovered\ relevant\ conditions}{Total\ number\ of\ ground\ truth\ conditions}\]
Precision refers to the ratio of discovered relevant conditions to the number of all discovered conditions.

\[Precision := \frac{Total\ number\ of\ discovered\ relevant\ conditions}{Total\ number\ of\ ground\ truth\ conditions}\]
The recall value can be used to evaluate if all underlying conditions have been discovered. Precision can be used to evaluate the complexity and comprehensibility of the discovered rules, i.e., if precision is $1$, only relevant conditions are extracted.
It is necessary to find a balance between these two values, as it is desirable to discover as many underlying conditions as possible without too many meaningless conditions which would add complexity and therefore lead to less comprehensible rules.
Therefore, the $F_1$ measure \cite{sammut_f1-measure_2010} can be used as it combines these recall and precision into one metric.
\[F_1 := 2*\frac{precision*recall}{precision+recall}\]
The closer to $1$, the higher are recall and precision and therefore the better is the ground truth represented by the discovered rule. Inversely, the closer to $0$ this value is, the worse it represents the underlying rule. The calculation of the $F_1$ measure is only possible if precision and recall are not zero, as otherwise an error will occur, due to division by 0. 
The comparison of the discovered conditions with the conditions contained in the ground truth rule is done manually. 

\subsection{Synthetic Data Set (Running Example)}
This data set consists of synthetic data for $2000$ workpieces based on the process model depicted in Fig. \ref{fig:process_before}. As result variable, we define the column 'res' that specifies if the workpiece is 'OK' or 'NOK'($1$ / $Branch1$ $\rightarrow$ OK, $0$ $\rightarrow$ NOK). The remaining variables are used as input columns. The ground truth is known for this case, as it was used to simulate the data and can be seen in \textbf{Ground Truth, Synthetic}.

\begin{tcolorbox}[center]\label{GT1}
{ \footnotesize \textbf{Ground Truth, Synthetic:}\\
$meas0>=range0$ AND $meas0<=range1$ AND $meas1>=range2$  AND $meas1<=range3$  AND $meas2>=range4$  AND $meas2<=range5$}
\end{tcolorbox}

The resulting \textbf{Rule BDT, Synthetic} (cf. next grey box) contains unary conditions and is discovered with an accuracy of 100\%. The discovered conditions almost match the specified ranges, but include some inaccuracies, e.g., $meas0>9.5$ implies that $9.6$ is already in the valid range which could lead to false results. As this approach is not able to correctly detect the binary conditions, recall and precision are $0$ and therefore the $F_1$ measure is undefined.

\begin{tcolorbox}[center]\label{BDT1}
{ \footnotesize \textbf{Rule BDT, Synthetic:} \\
WHEN $meas1>9.5$ AND $meas1<=20.5$ AND $meas2<=70.5$ AND $meas0>19.5$ AND $meas0<= 80.5$ AND $meas2>29.5$ THEN $Y=1$\\
Accuracy: $100\%$, \\
Recall: $0$,
Precision: $0$,
$F_1$: Undefined}
\end{tcolorbox}

BM discovers unary and binary conditions from the synthetic data set with an accuracy of $98.7\%$. However, only one of the binary conditions corresponds to the ground truth (cf. next grey box, marked in yellow). Accordingly, recall and precision are slightly higher than for BDT.  

\begin{tcolorbox}[center]
{ \footnotesize \textbf{Rule BM, Synthetic:}\\
WHEN $meas1<=range0$ AND $meas1<=range2$ AND $meas2<=70$ AND \colorbox{yellow}{$meas0<range1$} AND $meas0>=22$ AND $meas2 >=30$ THEN $Branch1$,\\
Accuracy: $98.7\%$, \\
Recall: $0.16$,
Precision: $0.16$,
$F_1$: $0.16$
}
\end{tcolorbox}

EDT discovers all applicable data conditions (even though in slightly different form, i.e., $meas0<range0 = false$ instead of $meas0>=range0$) achieving an accuracy of $100\%$ (cf. next grey box) and an $F_1$ measure of $1$. The resulting decision tree corresponds to Fig. \ref{tree:new}.

\begin{tcolorbox}[center]
{ \footnotesize \textbf{Rule EDT, Synthetic:} \\
WHEN $range2>meas1=false$ AND $meas1<=range3 = true$ AND $range5>=meas2 = true$ AND  $meas0<range0 = false$ AND $range1>=meas0 = true$ AND $range4>meas2 = false$ THEN $Y=1$, \\
Accuracy: $100\%$, \\
Recall: $1$,
Precision: $1$,
$F_1$: $1$}
\end{tcolorbox}

\subsection{BPIC 2017}
The BPIC 2017 data  
set\footnote{\url{https://doi.org/10.4121/uuid:5f3067df-f10b-45da-b98b-86ae4c7a310b}}
captures a loan application process. The result is defined as the decision if the loan conditions were accepted ($1$/$Branch1$ $\rightarrow$ accepted, $0$ $\rightarrow$ not accepted). As input all numerical variables are used as no prior information is available which variables are relevant for the decision rules. The ground truth decision rule is estimated, and will reflect the underlying rule only partly. However, we assume that that the offered amount being at least the same as the requested amount is a precondition for acceptance.

\begin{tcolorbox}[center]\label{GT2}
{ \footnotesize \textbf{Ground Truth, BPIC2017:}\\
$OfferedAmount>=RequestedAmount$}
\end{tcolorbox}

For BDT, the accuracy is $72\%$ and the conditions are meaningless with recall and precision values of $0$ and seem overly complex. For example the variable $NumberOfTerms$ is used several times and is unlikely to have an impact on the decision.

\begin{tcolorbox}[breakable, enhanced, center]
{ \footnotesize \textbf{Rule BDT, BPIC2017:}  \\
WHEN $CreditScore<=784.5$ AND $NumberOfTerms>47.5$ AND $Selected<= 0.5$ AND $FirstWithdrawalAmount<=9805.5$ AND $NumberOfTerms>55.5$ AND $FirstWithdrawalAmount<=3850.5$ AND $NumberOfTerms<=178.5$ AND $NumberOfTerms<=118.5$ AND $NumberOfTerms>75.5$ AND $RequestedAmount<= 57500.0$ AND $FirstWithdrawalAmount<= 2075.19$ AND $FirstWithdrawalAmount<=2.5$ AND $NumberOfTerms>78.5$ AND $NumberOfTerms>110.5$ AND $NumberOfTerms<= 113.5$ AND $RequestedAmount>3119.5$ THEN $Y=1$, \\Accuracy: $72\%$, \\
Recall: $0$,
Precision: $0$,
$F_1$: Undefined}
\end{tcolorbox}
 
The following rule \textbf{Rule BM, BPIC2017 } detected by BM achieves an accuracy of $70\%$. Its semantics is unclear and unintuitive as variables \texttt{CreditScore} and \texttt{MonthlyCosts} seem unrelated. This is reflected in the recall and precision values.

\begin{tcolorbox}[center]
{ \footnotesize \textbf{Rule BM, BPIC2017:} \\
WHEN $CreditScore!=MonthlyCost$ THEN $Branch1$, \\Accuracy: $70\%$, \\
Recall: $0$,
Precision: $0$,
$F_1$: Undefined}
\end{tcolorbox}

EDT discovers rule \textbf{Rule EDT, BPIC2017} that is similarly complex to \textbf{Rule BDT, BPIC2017}. It includes at least one binary condition that seems relevant, i.e., $RequestedAmount<=OfferedAmount$ as highlighted in yellow. The rule also includes apparently meaningless conditions such as \textsl{$MonthlyCosts >= NumberOfTerms$}. \footnote{The corresponding decision tree yields no further insights and is therefore and due to lack of space not displayed here.} The accuracy is $73$\% and slightly exceeds the ones of BDT and BM for BPIC2017. The $F_1$ measure is $0.25$, due to a recall value of $1$ and the precision value of $0.14$. However, the value of $1$ as recall value is most likely an overestimation as only one underlying condition was defined, whereas the true ground truth probably contains more conditions. 

\begin{tcolorbox}[center]
{ \footnotesize \textbf{Rule EDT, BPIC2017:}  \\
WHEN $NumberOfTerms<=CreditScore = false$ AND $Selected<=CreditScore = true$ AND \colorbox{yellow}{$RequestedAmount<=OfferedAmount = true$} AND $FirstWithdrawalAmount<=CreditScore = false$ AND $MonthlyCost>=NumberOfTerms = true$ AND $MonthlyCost<=FirstWithdrawalAmount = true$  AND $FirstWithdrawalAmount>=OfferedAmount = true$ THEN $Y=1$, \\Accuracy: $73\%$, \\
Recall: $1$,
Precision: $0.14$,
$F_1$: $0.25$}
\end{tcolorbox}

\subsection{BPIC 2020}
The BPIC2020 dataset\footnote{\url{https://doi.org/10.4121/uuid:52fb97d4-4588-43c9-9d04-3604d4613b51}} contains the process log of a travel permit and expense system. We define the column 'Overspent', which specifies if more money was spend than initially authorized ($1$ / $Branch1$ $\rightarrow$ overspent, $0$ $\rightarrow$ not overspent) as result variable, which also leads to the definition of the ground truth. All other numerical variables serve as input. 

\begin{tcolorbox}[center]\label{GT3}
{ \footnotesize \textbf{Ground Truth, BPIC2020:}\\
$RequestedBudget<TotalDeclared$}
\end{tcolorbox}

For BDT, the accuracy is $100\%$ and one unary condition in \textbf{Rule BDT, BPIC2020} is discovered, which seems meaningless, also being reflected in the recall and precision values. We would rather expect $OverspentAmount>0$ as rule, because as soon as anything more than agreed upon was spent, it would be 'overspent'.

\begin{tcolorbox}[center]
{ \footnotesize \textbf{Rule BDT, BPIC2020:} \\
WHEN $OverspentAmount>10.11$ THEN $Y=1$,\\ Accuracy: $100\%$, \\
Recall: $0$,
Precision: $0$,
$F_1$: Undefined}
\end{tcolorbox}

For BM, one binary condition in \textbf{Rule BM, BPIC2020} is discovered, saying that if the amount that was overspent is higher than the total requested budget it was 'overspent'. The accuracy is $84.5\%$. This most likely does not reflect the underlying rules. Therefore recall and precision are $0$.

\begin{tcolorbox}[center]
{ \footnotesize \textbf{Rule BM, BPIC2020:} \\
WHEN $!(OverspentAmount$ $>$ $RequestedBudget)$ THEN $Branch1$, \\Accuracy: $84.5\%$, \\
Recall: $0$,
Precision: $0$,
$F_1$: Undefined}
\end{tcolorbox}

\begin{figure}[htb!]
 \centering 
\begin{forest} 
[$RequestedBuget<TotalDeclared$
    [Not Overspent, edge label={node[midway,left]{false}}]   
    [Overspent, edge label={node[midway,right]{true}}]   
] 
\end{forest}
\caption{Decision tree generated by EDT for BPI20.}
\label{tree:bpi}
\vspace{-1mm}
\end{figure}
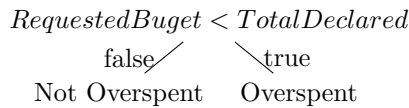

The accuracy of $83\%$ for EDT is below the accuracy values obtained for BDT and BM on BPIC2020. However, the discovered rule \textbf{Rule EDT, BPIC2020} seems more meaningful: it states that as soon as the requested budget is smaller than the totally declared one, i.e., the money that was spent, overspent is true, leading to an overall $F_1$ value of $1$. The resulting decision tree can be seen in Fig. \ref{tree:bpi}

\begin{tcolorbox}[center]
{ \footnotesize \textbf{Rule EDT, BPIC2020:}  \\
WHEN $RequestedBudget<TotalDeclared = true$ THEN $Y=1$, \\Accuracy: $83\%$, \\
Recall: $1$,
Precision: $1$,
$F_1$: $1$}
\end{tcolorbox}

\section{Discussion}
\label{sect:discussion}

The ground truth is only known for the running example. The results for the other two datasets are interpreted using the available description. All results can be seen in Tab. \ref{eval_results}.

For the running example, BDT extracts unary conditions similar to the dimension ranges, leading to an accuracy of $100\%$. However, the accuracy can drop if the ranges change or if the measured values lie closer to the ends of the ranges. BM detects unary and binary conditions, with an accuracy of $98.7\%$. The discovered rule partly reflects the underlying rule. EDT discovered all underlying conditions and achieves an accuracy of $100\%$. It is interesting that EDT achieves a more meaningful result than BM, even though it is based on BM. The more meaningful results also reflects in a higher $F_1$ value, which is $1$ for EDT and $0.16$ for BM. 

For the BPIC17, the accuracy for all approaches is around $70\%$. BDT leads to an extensive rule, however the discovered conditions seem meaningless. BM detects a single binary condition, which also contains no semantic meaningfulness. EDT generates an extensive rule. However, only one of the discovered conditions is meaningful. With this exception, none of the approaches was able to detect meaningful decision rules, which is also reflected in the corresponding $F_1$ values. This can be an indication that the discovery of the underlying rules is not possible with the tested approaches or that is is not possible at all, i.e., the decision depends on additional factors that are not part of the log files, e.g. a loan offer from a different bank.

For the BPIC20, BDT and BM each discover one condition that does not contain meaningful content, leading to an undefined $F_1$ value. In contrast, EDT generates one binary condition that is meaningful, i.e., it was overspent if the requested budget is smaller than the actual spent amount, which is also reflected in the $F_1$ value. BM and EDT achieve an accuracy around $80\%$. Interestingly, the highest accuracy was achieved by BDT. This is possible due to overfitting of the data, i.e., only cases where the  $OverspentAmount$ was greater than $10.11$ are part of the dataset. Overfitting can be partly prevented by splitting training and test set, however in this case it was not possible to avoid overfitting. 

\begin{table}[]
\caption{Results.}
\label{eval_results}
\begin{tabular}{lllll}
\textbf{Rule} & \textbf{Accuracy} & \textbf{Recall} & \textbf{Precision} & \textbf{F1 Measure} \\
BDT, Synthetic & 100\% & 0    & 0    & U*  \\
BM, Synthetic  & 98\%  & 0.16 & 0.16 & 0.16 \\
EDT, Synthetic & 100\% & 1    & 1    & 1    \\
BDT, BPIC2017  & 72\%  & 0    & 0    & U   \\
BM, BPIC2017   & 70\%  & 0    & 0    & U   \\
EDT, BPIC2017  & 73\%  & 1    & 0.14 & 0.25 \\
BDT, BPIC2020  & 100\% & 0    & 0    & U   \\
BM, BPIC2020   & 84\%  & 0    & 0    & U   \\
EDT, BPIC2020  & 83\%  & 1    & 1    & 1    \\
               &       &      &      &      \\
\multicolumn{5}{l}{*U - Undefined, due to division by $0$}                                
\end{tabular}
\end{table}

From the evaluation some statements can be derived:
Firstly, the accuracy of decision rules can be high, up to $100\%$, even though the discovered rules do not display the underlying decision logic. Conversely, a lower accuracy does not mean that the rules do not contain meaningful decision logic. 
If the ground truth is available, recall and precision and therefore the $F_1$ value can be used to evaluate the meaningfulness of the extracted rules.
Secondly, the ground truth can be a combination of unary and binary data conditions, or cannot be discovered, because the conditions are more complex or the deciding factors are not part of the event log. Thirdly, the data quality is important. This includes completeness, i.e., are all possible values represented by the training data, as well as noise, i.e., the more noise the data contains the more difficult it is to find meaningful rules. And fourth, BM and EDT discover different conditions, even though they build upon the same principle of creating latent variables. A possible explanation could be the different ways of choosing variables to include in the decision tree. BM preselects variables using information gain as metric, whereas in EDT all variables are used as input for the decision tree algorithm which is responsible for choosing the ones containing the most information. Another possibility is, that the different decision tree implementations lead to different results. Interestingly, BM would be able to detect more complex relations, but no linear relationships or relations using arithmetic operators were discovered. This leads to the assumption that such relations are not part of the used datasets and BM may yield better results if such relationships are contained. The additional decision trees in EDT can provide added value and more insight on when a specific data condition is used.
Therefore a complementary approach, i.e. a combination of multiple approaches may be able to generate the most accurate and meaningful rules.

The following recommendations can support the discovery of meaningful and accurate decision rules.
\begin{itemize}
    \item The possibility of binary data conditions should be taken into account, especially in regards to choosing a decision mining algorithm.
    \item Multiple decision mining algorithms can be used in conjunction to discover potential rules.
    \item The discovered decision rules can be seen as \texttt{candidate rules}.
    \item A domain expert should be able to identify meaningful decision rules from the candidate rules.
    \item A comparison between possible values and actual values in the dataset can yield more insight into the decision logic.
    \item The possibility that no meaningful decision rules can be discovered should be kept in mind.
    \item If the ground truth is available, recall and precision and therefore the $F_1$ value can be use to evaluate the meaningfulness of the extracted rules. 
\end{itemize}
However, many challenges with regard to semantics and meaningfulness of decision rules remain.

\subsection{Future Challenges}
Some variables are meaningful as part of unary conditions, e.g. 'Monthly Cost' in BPIC17, other variables only in relation to other variables, e.g., measurements in the running example. Determining if a variable should be best used in a unary or in a binary condition could simplify the discovery. Generally, developing a metric to specify the meaningfulness of a decision rule if no underlying rules are known, could be used, in conjunction with the accuracy, to find the most appropriate rules. Furthermore, the integration of more complex decision logic should be explored, e.g. time-series data. The running example also contained some other challenges in process mining, e.g. the information contained in the engineering drawing has to be converted into a more structured format before it is included as part of the event log. Additionally, the current approaches cannot include lists as input format. Future challenges therefore remain to integrate non-structured variables. 

\subsection{Limitations and threats to validity}
This study is limited to three datasets, one of them consisting of synthetic data. Therefore, the ability of the approaches to produce meaningful rules cannot be generalized. However, it shows that a high accuracy is not equivalent to meaningfulness. Furthermore, not all complex rules are covered by the used approaches, e.g. time series data. In addition, only numerical data variables are taken into account. These limitations will be part of future work.

\section{Related Work}
\label{sect:rw}
\cite{rozinat_decision_2006} introduced the term Decision Mining (DM), by describing an approach to discover the different paths that can be taken in a process. \cite{leoni_decision_2018} give an overview of the state-of-the-art where DM is seen as a classification problem and typically employs decision trees as these are able to generate human readable rules. \cite{de_leoni_data-aware_2013} focus on the data-dependency of decisions, investigating the relationship between the data-flow and the taken decisions. \cite{mannhardt_decision_2016} describe an approach that is able to work with overlapping rules that are beneficiary if data is missing or the underlying rules are not fully deterministic. \cite{mannhardt_data-driven_2017} deal with infrequently used paths that would often be disregarded as noise, but can reveal interesting conditions that rarely evaluate to true.  \cite{dunkl_method_2015} present an approach to discover rules that involve time-series data, e.g., if the temperature is above a certain threshold for a certain time. 
However, all of these approaches only take unary conditions into account.
To the best of our knowledge, BranchMiner \cite{de_leoni_discovering_2013} is the first approach focusing on binary data conditions. \cite{leno_automated_2020,maggi_discovering_2013} deal with discovering Declare constraints that also involve binary data conditions. \cite{bazhenova_discovering_2016} builds on the work of \cite{de_leoni_discovering_2013}, but in the context of creating DMN \cite{group_omg_decision_2020} models from event logs. 

This paper is complementary to the above mentioned works in that it builds upon existing approaches and evaluates them on real life datasets in order to differentiate between accuracy and semantically meaningful rules.  

\section{Conclusion}
\label{sect:conclusion}

In this paper, we compared three decision mining approaches with regards to their ability to detect meaningful decision rules from event logs using three different datasets, two containing real life data. The evaluation of decision rules with regards to semantics and meaningfulness has not been an area of focus so far. The results show that each of the three approaches discovered accurate decision rules, but only some contained meaningful decision logic. One main observation is that a high accuracy does not guarantee meaningful rules and additional measures have to be taken. 
We stated some guidelines to improve meaningfulness. This can be done using different decision mining approaches to generate a set of \texttt{candidate rules}, which are then evaluated by a domain expert.
In future work, we will evaluate decision mining results with domain experts and develop metrics for meaningfulness if the underlying rules are not known, as complement to accuracy. 

Overall, this study provides a first step towards explainable decision mining through understandable decision mining results.

\section*{Acknowledgment}
This work has been partially supported and funded by the Austrian Research Promotion Agency (FFG) via the Austrian Competence Center for Digital Production (CDP) under the contract number 881843.
\bibliographystyle{IEEEtran}
\bibliography{extracted_full}

\begin{thebibliography}{10}
\providecommand{\url}[1]{#1}
\csname url@samestyle\endcsname
\providecommand{\newblock}{\relax}
\providecommand{\bibinfo}[2]{#2}
\providecommand{\BIBentrySTDinterwordspacing}{\spaceskip=0pt\relax}
\providecommand{\BIBentryALTinterwordstretchfactor}{4}
\providecommand{\BIBentryALTinterwordspacing}{\spaceskip=\fontdimen2\font plus
\BIBentryALTinterwordstretchfactor\fontdimen3\font minus
  \fontdimen4\font\relax}
\providecommand{\BIBforeignlanguage}[2]{{%
\expandafter\ifx\csname l@#1\endcsname\relax
\typeout{** WARNING: IEEEtran.bst: No hyphenation pattern has been}%
\typeout{** loaded for the language `#1'. Using the pattern for}%
\typeout{** the default language instead.}%
\else
\language=\csname l@#1\endcsname
\fi
#2}}
\providecommand{\BIBdecl}{\relax}
\BIBdecl

\bibitem{aalst_process_2016}
\BIBentryALTinterwordspacing
W.~van~der Aalst, ``\BIBforeignlanguage{en}{Data {Science} in {Action}},'' in
  \emph{\BIBforeignlanguage{en}{Process {Mining}: {Data} {Science} in
  {Action}}}, W.~van~der Aalst, Ed.\hskip 1em plus 0.5em minus 0.4em\relax
  Berlin, Heidelberg: Springer, 2016, pp. 3--23. [Online]. Available:
  \url{https://doi.org/10.1007/978-3-662-49851-4_1}
\BIBentrySTDinterwordspacing

\bibitem{leoni_decision_2018}
\BIBentryALTinterwordspacing
M.~d. Leoni and F.~Mannhardt, ``\BIBforeignlanguage{en}{Decision {Discovery} in
  {Business} {Processes}},'' in \emph{\BIBforeignlanguage{en}{Encyclopedia of
  {Big} {Data} {Technologies}}}, S.~Sakr and A.~Zomaya, Eds.\hskip 1em plus
  0.5em minus 0.4em\relax Cham: Springer International Publishing, 2018, pp.
  1--12. [Online]. Available:
  \url{http://link.springer.com/10.1007/978-3-319-63962-8_96-1}
\BIBentrySTDinterwordspacing

\bibitem{dunkl_method_2015}
R.~Dunkl, S.~Rinderle-Ma, W.~Grossmann, and K.~Anton~Fröschl,
  ``\BIBforeignlanguage{en}{A {Method} for {Analyzing} {Time} {Series} {Data}
  in {Process} {Mining}: {Application} and {Extension} of {Decision} {Point}
  {Analysis}},'' in \emph{\BIBforeignlanguage{en}{Information {Systems}
  {Engineering} in {Complex} {Environments}}}, ser. Lecture {Notes} in
  {Business} {Information} {Processing}, S.~Nurcan and E.~Pimenidis, Eds.\hskip
  1em plus 0.5em minus 0.4em\relax Cham: Springer International Publishing,
  2015, pp. 68--84.

\bibitem{de_leoni_discovering_2013}
M.~de~Leoni, M.~Dumas, and L.~García-Bañuelos,
  ``\BIBforeignlanguage{en}{Discovering {Branching} {Conditions} from
  {Business} {Process} {Execution} {Logs}},'' in
  \emph{\BIBforeignlanguage{en}{Fundamental {Approaches} to {Software}
  {Engineering}}}, ser. Lecture {Notes} in {Computer} {Science}, V.~Cortellessa
  and D.~Varró, Eds.\hskip 1em plus 0.5em minus 0.4em\relax Berlin,
  Heidelberg: Springer, 2013, pp. 114--129.

\bibitem{leewis_future_2020}
S.~Leewis, M.~Berkhout, and K.~Smit, ``Future {Challenges} in {Decision}
  {Mining} at {Governmental} {Institutions},'' in \emph{{AMCIS} 2020
  {Proceedings}}, Aug. 2020.

\bibitem{guidotti_survey_2018}
\BIBentryALTinterwordspacing
R.~Guidotti, A.~Monreale, S.~Ruggieri, F.~Turini, F.~Giannotti, and
  D.~Pedreschi, ``A {Survey} of {Methods} for {Explaining} {Black} {Box}
  {Models},'' \emph{ACM Comput. Surv.}, vol.~51, no.~5, pp. 93:1--93:42, Aug.
  2018. [Online]. Available: \url{http://doi.org/10.1145/3236009}
\BIBentrySTDinterwordspacing

\bibitem{breiman_classification_1984}
L.~Breiman, J.~Friedman, C.~J. Stone, and R.~A. Olshen,
  \emph{\BIBforeignlanguage{en}{Classification and {Regression}
  {Trees}}}.\hskip 1em plus 0.5em minus 0.4em\relax Taylor \& Francis, Jan.
  1984.

\bibitem{quinlan_j_r_c45_1993}
{Quinlan, J. R.}, \emph{C4.5: {Programs} for {Machine} {Learning}}.\hskip 1em
  plus 0.5em minus 0.4em\relax Morgan Kaufmann Publishers, 1993.

\bibitem{pedregosa_scikit-learn_2011}
F.~Pedregosa, G.~Varoquaux, A.~Gramfort, V.~Michel, B.~Thirion, O.~Grisel,
  M.~Blondel, P.~Prettenhofer, R.~Weiss, V.~Dubourg, J.~Vanderplas, A.~Passos,
  D.~Cournapeau, M.~Brucher, M.~Perrot, and E.~Duchesnay, ``Scikit-learn:
  {Machine} {Learning} in {Python},'' \emph{Journal of Machine Learning
  Research}, vol.~12, pp. 2825--2830, 2011.

\bibitem{ernst_daikon_2007}
\BIBentryALTinterwordspacing
M.~D. Ernst, J.~H. Perkins, P.~J. Guo, S.~McCamant, C.~Pacheco, M.~S. Tschantz,
  and C.~Xiao, ``\BIBforeignlanguage{en}{The {Daikon} system for dynamic
  detection of likely invariants},'' \emph{\BIBforeignlanguage{en}{Science of
  Computer Programming}}, vol.~69, no.~1, pp. 35--45, Dec. 2007. [Online].
  Available: \url{https://www.sciencedirect.com/science/article/pii/
  S016764230700161X}
\BIBentrySTDinterwordspacing

\bibitem{eibe_weka_2016}
F.~Eibe, M.~A. Hall, and {Witten, Ian H.}, \emph{The {WEKA} {Workbench}.
  {Online} {Appendix} for "{Data} {Mining}: {Practical} {Machine} {Learning}
  {Tools} and {Techniques}"}, 4th~ed.\hskip 1em plus 0.5em minus 0.4em\relax
  Morgan Kaufman, 2016.

\bibitem{blanco-justicia_machine_2020}
\BIBentryALTinterwordspacing
A.~Blanco-Justicia, J.~Domingo-Ferrer, S.~Martínez, and D.~Sánchez,
  ``\BIBforeignlanguage{en}{Machine learning explainability via
  microaggregation and shallow decision trees},''
  \emph{\BIBforeignlanguage{en}{Knowledge-Based Systems}}, vol. 194, p. 105532,
  Apr. 2020. [Online]. Available:
  \url{https://www.sciencedirect.com/science/article/pii/S0950705120300368}
\BIBentrySTDinterwordspacing

\bibitem{ting_precision_2010}
\BIBentryALTinterwordspacing
K.~M. Ting, ``Precision and {Recall},'' in \emph{Encyclopedia of {Machine}
  {Learning}}, C.~Sammut and G.~I. Webb, Eds.\hskip 1em plus 0.5em minus
  0.4em\relax Boston, MA: Springer US, 2010, p. 781. [Online]. Available:
  \url{https://doi.org/10.1007/978-0-387-30164-8_652}
\BIBentrySTDinterwordspacing

\bibitem{sammut_f1-measure_2010}
\BIBentryALTinterwordspacing
``F1-{Measure},'' in \emph{Encyclopedia of {Machine} {Learning}}, C.~Sammut and
  G.~I. Webb, Eds.\hskip 1em plus 0.5em minus 0.4em\relax Boston, MA: Springer
  US, 2010, pp. 397--397. [Online]. Available:
  \url{https://doi.org/10.1007/978-0-387-30164-8_298}
\BIBentrySTDinterwordspacing

\bibitem{rozinat_decision_2006}
A.~Rozinat and W.~M.~P. van~der Aalst, ``\BIBforeignlanguage{en}{Decision
  {Mining} in {ProM}},'' in \emph{\BIBforeignlanguage{en}{Business {Process}
  {Management}}}, ser. Lecture {Notes} in {Computer} {Science}, S.~Dustdar,
  J.~L. Fiadeiro, and A.~P. Sheth, Eds.\hskip 1em plus 0.5em minus 0.4em\relax
  Berlin, Heidelberg: Springer, 2006, pp. 420--425.

\bibitem{de_leoni_data-aware_2013}
\BIBentryALTinterwordspacing
M.~de~Leoni and W.~M.~P. van~der Aalst, ``Data-aware process mining:
  discovering decisions in processes using alignments,'' in \emph{Proceedings
  of the 28th {Annual} {ACM} {Symposium} on {Applied} {Computing}}, ser. {SAC}
  '13.\hskip 1em plus 0.5em minus 0.4em\relax New York, NY, USA: Association
  for Computing Machinery, Mar. 2013, pp. 1454--1461. [Online]. Available:
  \url{https://doi.org/10.1145/2480362.2480633}
\BIBentrySTDinterwordspacing

\bibitem{mannhardt_decision_2016}
\BIBentryALTinterwordspacing
F.~Mannhardt, M.~d. Leoni, H.~A. Reijers, and W.~M. P. v.~d. Aalst,
  ``\BIBforeignlanguage{en}{Decision {Mining} {Revisited} - {Discovering}
  {Overlapping} {Rules}},'' in \emph{\BIBforeignlanguage{en}{Advanced
  {Information} {Systems} {Engineering}}}.\hskip 1em plus 0.5em minus
  0.4em\relax Cham: Springer, Jun. 2016, pp. 377--392. [Online]. Available:
  \url{https://link.springer.com/chapter/10.1007/978-3-319-39696- 5_23}
\BIBentrySTDinterwordspacing

\bibitem{mannhardt_data-driven_2017}
F.~Mannhardt, M.~de~Leoni, H.~A. Reijers, and W.~M.~P. van~der Aalst,
  ``\BIBforeignlanguage{en}{Data-{Driven} {Process} {Discovery} - {Revealing}
  {Conditional} {Infrequent} {Behavior} from {Event} {Logs}},'' in
  \emph{\BIBforeignlanguage{en}{Advanced {Information} {Systems}
  {Engineering}}}, ser. Lecture {Notes} in {Computer} {Science}, E.~Dubois and
  K.~Pohl, Eds.\hskip 1em plus 0.5em minus 0.4em\relax Cham: Springer
  International Publishing, 2017, pp. 545--560.

\bibitem{leno_automated_2020}
\BIBentryALTinterwordspacing
V.~Leno, M.~Dumas, F.~M. Maggi, M.~La~Rosa, and A.~Polyvyanyy,
  ``\BIBforeignlanguage{en}{Automated discovery of declarative process models
  with correlated data conditions},'' \emph{\BIBforeignlanguage{en}{Information
  Systems}}, vol.~89, p. 101482, Mar. 2020. [Online]. Available:
  \url{https://www.sciencedirect.com/science/article/pii/S0306437919305344}
\BIBentrySTDinterwordspacing

\bibitem{maggi_discovering_2013}
F.~M. Maggi, M.~Dumas, L.~García-Bañuelos, and M.~Montali,
  ``\BIBforeignlanguage{en}{Discovering {Data}-{Aware} {Declarative} {Process}
  {Models} from {Event} {Logs}},'' in \emph{\BIBforeignlanguage{en}{Business
  {Process} {Management}}}, ser. Lecture {Notes} in {Computer} {Science},
  F.~Daniel, J.~Wang, and B.~Weber, Eds.\hskip 1em plus 0.5em minus 0.4em\relax
  Berlin, Heidelberg: Springer, 2013, pp. 81--96.

\bibitem{bazhenova_discovering_2016}
E.~Bazhenova, S.~Buelow, and M.~Weske, ``\BIBforeignlanguage{en}{Discovering
  {Decision} {Models} from {Event} {Logs}},'' in
  \emph{\BIBforeignlanguage{en}{Business {Information} {Systems}}}, ser.
  Lecture {Notes} in {Business} {Information} {Processing}, W.~Abramowicz,
  R.~Alt, and B.~Franczyk, Eds.\hskip 1em plus 0.5em minus 0.4em\relax Cham:
  Springer International Publishing, 2016, pp. 237--251.

\bibitem{group_omg_decision_2020}
\BIBentryALTinterwordspacing
O.~M. Group~(OMG), ``\BIBforeignlanguage{en}{Decision {Model} and {Notation} -
  {DMN}},'' 2020. [Online]. Available: \url{https://www.omg.org/spec/DMN/1.3/}
\BIBentrySTDinterwordspacing

\end{thebibliography}

\end{document}